\documentclass[letterpaper, 10 pt, conference]{ieeeconf}

\usepackage{amsmath,amsfonts}
\usepackage{algorithmic}
\usepackage{algorithm}
\usepackage{array}
\usepackage[caption=false,font=normalsize,labelfont=sf,textfont=sf]{subfig}
\usepackage{textcomp}
\usepackage{stfloats}
\usepackage{url}
\usepackage{verbatim}
\usepackage{graphicx}
\usepackage{cite}
\usepackage{hyperref}
\usepackage{makecell}

\begin{document}

\title{Pedestrian motion prediction evaluation for urban autonomous driving}

\author{
\authorblockN{Dmytro Zabolotnii}
\authorblockA{Institute of Computer Science, \\ University of Tartu, \\ Narva mnt 18, Tartu 51009, Estonia. \\ Email: dmytro.zabolotnii@ut.ee. \\ Corresponding author.}
\and
\authorblockN{Yar Muhammad}
\authorblockA{Department of Computer Science, \\ University of Hertfordshire, \\ AL10 9AB Hatfield, U.K. \\ Email: y.muhammad@herts.ac.uk}
\and
\authorblockN{Naveed Muhammad}
\authorblockA{Institute of Computer Science, \\ University of Tartu, \\ Narva mnt 18, Tartu 51009, Estonia. \\ Email: naveed.muhammad@ut.ee}
}

% \author{Dmytro Zabolotnii$^{1}$, Yar Muhammad$^{2}$, Naveed Muhammad$^{3}$
%         % <-this % stops a space
% \thanks{Manuscript received: Month, Day, Year; Revised Month, Day, Year; Accepted Month, Day, Year.}%Use only for final RAL version
% \thanks{This paper was recommended for publication by Editor Editor A. Name upon evaluation of the Associate Editor and Reviewers' comments.
% This work was supported by (organizations/grants which supported the work.)} %Use only for final RAL version
% \thanks{$^{1}$ D. Zabolotnii is with Institute of Computer Science, University of Tartu, Narva mnt 18, Tartu 51009, Estonia. {\tt\footnotesize dmytro.zabolotnii@ut.ee.} Corresponding author.} % <-this % stops a space
% \thanks{$^{2}$ Y. Muhammad is with Department of Computer Science, School of Physics, Engineering and Computer Science, University of Hertfordshire, AL10 9AB Hatfield, U.K. {\tt\footnotesize y.muhammad@herts.ac.uk}} % <-this % stops a space
% \thanks{$^{3}$ N. Muhammad is with Institute of Computer Science, University of Tartu, Narva mnt 18, Tartu 51009, Estonia. {\tt\footnotesize naveed.muhammad@ut.ee.}} 
% \thanks{Digital Object Identifier (DOI): see top of this page.}
% }

% % The paper headers
% \markboth{IEEE Robotics and Automation Letters. Preprint Version. Accepted \textbf{Month}, \textbf{Year}}
% {Zabolotnii \MakeLowercase{\textit{et al.}}: Prediction Evaluation} 

% \IEEEpubid{0000--0000/00\$00.00~\copyright~2024 IEEE}
% Remember, if you use this you must call \IEEEpubidadjcol in the second
% column for its text to clear the IEEEpubid mark.

\maketitle

\begin{abstract}
Pedestrian motion prediction is a key part of the modular-based autonomous driving pipeline, ensuring safe, accurate, and timely awareness of human agents' possible future trajectories. The autonomous vehicle can use this information to prevent any possible accidents and create a comfortable and pleasant driving experience for the passengers and pedestrians. A wealth of research was done on the topic from the authors of robotics, computer vision, intelligent transportation systems, and other fields. However, a relatively unexplored angle is the integration of the state-of-art solutions into existing autonomous driving stacks and evaluating them in real-life conditions rather than sanitized datasets. We analyze selected publications with provided open-source solutions and provide a perspective obtained by integrating them into existing Autonomous Driving framework - Autoware Mini and performing experiments in natural urban conditions in Tartu, Estonia to determine valuability of traditional motion prediction metrics. This perspective should be valuable to any potential autonomous driving or robotics engineer looking for the real-world performance of the existing state-of-art pedestrian motion prediction problem. The code with instructions on accessing the dataset is available at \href{https://github.com/dmytrozabolotnii/autoware_mini}{https://github.com/dmytrozabolotnii/autoware\_mini}.

Index Terms-AI-Enabled Robotics, Autonomous Vehicle Navigation, Computer Vision for Transportation, Datasets for Human Motion, Human Detection and Tracking, Intention Recognition

\end{abstract}

% \begin{IEEEkeywords}
% AI-Enabled Robotics, Autonomous Vehicle Navigation, Computer Vision for Transportation, Datasets for Human Motion, Human Detection and Tracking, Intention Recognition
% \end{IEEEkeywords}

\section{Introduction}

Autonomous driving research is an exciting topic nowadays, with an aggrandizing promise to improve the driving process for everyone and eventually replace human drivers, starting from future long-distance truck operators \cite{herrmann2018autonomous} to already running autonomous taxi services \cite{californiadmv_2021}. However, even if the human drivers can be replaced, the human factor will never disappear from autonomous driving. One of the most significant human factors is not even connected to the car itself - rather to the agency of other human actors on the road - other drivers in non-self-driving cars and pedestrians \cite{rudenko2020human}. A self-driving vehicle should be aware of their possible future actions and apply necessary corrections to its course of action to avoid collisions, follow legal directives, and ensure a safe and comfortable environment for all actors. This function is essential for ensuring pedestrian safety. Poor interactions of the vehicle and pedestrians lead to many traffic accidents, with over 80\% possibility of the fatal outcome when the car moves over 60 km/h \cite{zegeer2012pedestrian}. This issue is even more exorbitant among youth and elderly groups, especially in developing countries, with road-caused accidents being the lead cause of youth disabilities globally \cite{branche2008world}.

% \IEEEPARstart{A}{utonomous} driving research is an exciting topic nowadays, with an aggrandizing promise to improve the driving process for everyone and eventually replace human drivers, starting from future long-distance truck operators \cite{herrmann2018autonomous} to already running autonomous taxi services \cite{californiadmv_2021}. However, even if the human drivers can be replaced, the human factor will never disappear from autonomous driving. One of the most significant human factors is not even connected to the car itself - rather to the agency of other human actors on the road - other drivers in non-self-driving cars and pedestrians \cite{rudenko2020human}. A self-driving vehicle should be aware of their possible future actions and apply necessary corrections to its course of action to avoid collisions, follow legal directives, and ensure a safe and comfortable environment for all actors. This function is essential for ensuring pedestrian safety. Poor interactions of the vehicle and pedestrians lead to many traffic accidents, with over 80\% possibility of the fatal outcome when the car moves over 60 km/h \cite{zegeer2012pedestrian}. This issue is even more exorbitant among youth and elderly groups, especially in developing countries, with road-caused accidents being the lead cause of youth disabilities globally \cite{branche2008world}.

Accordingly, to solve these problems in the future autonomous driving applications, pedestrian motion prediction remains an active research topic, with numerous new solutions published every year. Starting from the classic physics-based models \cite{scholler2020constant}, the trends shifted to Machine-Learning-based solutions over the last decade \cite{rudenko2020human}. In the last few years of the published articles, there is a wide representation of underlying architectures used for prediction \cite{schuetz2023review}: CVAE methods \cite{lee2022muse, zhou2023csr, wang2022stepwise, chen2022scept}, GAN methods \cite{li2021temporal, zhou2022gchgat, yang2023social}, Transformer methods \cite{ngiam2021scene, nayakanti2023wayformer, cheng2023gatraj}, Diffusion methods \cite{mao2023leapfrog, bae2024singulartrajectory}, and even as novel application of LLM-based methods \cite{keysan2023can, bae2024can}. However, while the variety of underlying methods is excellent, the evaluation of these methods is not as outstanding. The majority of newly published (that often claim to be state-of-the-art), use ADE/FDE (or minADE/minFDE variants) metrics for validation, despite emerging research about deficiencies of these metrics \cite{ivanovic2021rethinking, wu2023truly, shridhar2021beelines}  and the existance of appropriate alternative metrics that have been proposed in literature, such as negative log-likelihood (NLL) or Average
Mahalanobis Distance (AMD) \cite{mohamed2022social}. Additionally, model training and experimental evaluation are often done only on small and somewhat outdated datasets such as ETH/UCY \cite{pellegrini2010improving, lerner2007crowds} and SDD \cite{robicquet2016learning} instead of larger and much more diverse datasets such as Argoverse \cite{chang2019argoverse} or Waymo Open Motion \cite{ettinger2021large}. 

\begin{figure}[!t]
\centering
\includegraphics[width=3.45in]{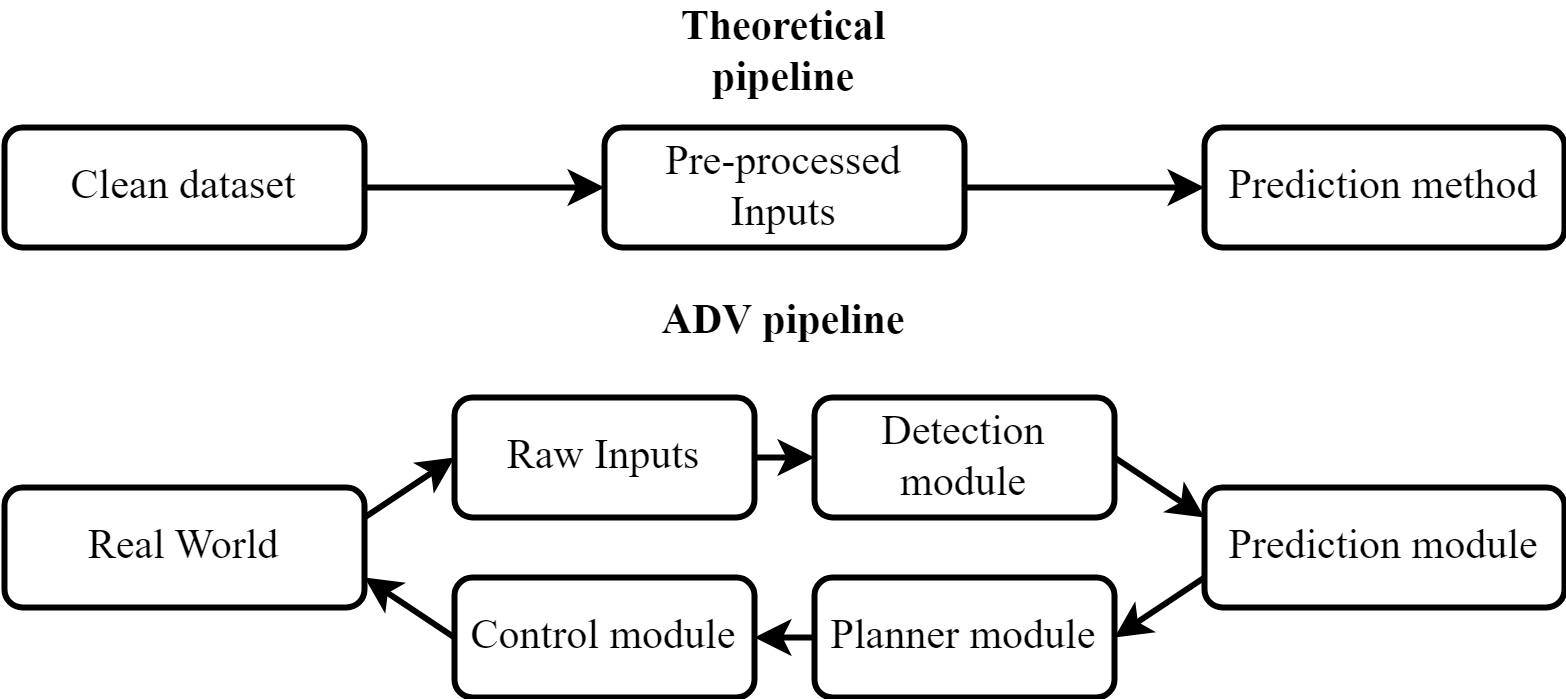}
\caption{Architectural difference between implementation of state-of-art motion prediction methods and their potential implementation in a modular Autonomous Driving framework.}
\label{fig_architecture_gap}
\end{figure}

Even more fundamentally, there remains an inconsistency between evaluating any prediction method on the offline dataset and evaluating the prediction method as the module in the modular Autonomous Driving framework, as illustrated by Fig. \ref{fig_architecture_gap}. During the application of the prediction algorithm as part of the modular framework, not only is it affected by the potential errors in the upstream detection and sensor modules, but it also influences the planning and control modules that lead to the movement of the ego-vehicle, and correspondingly all the exo-agents whose motion is predicted can change their future actions because of our movements. This compounding error effect leads to what is called \emph{dynamics gap} in \cite{wu2023truly}. Combining all these factors demonstrates the need for a change in the approach to pedestrian motion prediction evaluation. 

In this letter, we aim to expand upon the evaluation of existing state-of-the-art methods using a more realistic experimental setup - an open-source autonomous driving framework with real-life sensor data. The necessity of processing raw sensor data and running all the other parts of the autonomous driving framework simultaneously as the prediction algorithms on the limited consumer-grade hardware that is expected to be in the autonomous vehicle requires stringent evaluation of the computational efficiency. At the same time, we want to challenge the standard Best-of-\emph{N} approach that uses ADE/FDE metrics for evaluating the prediction algorithm in the existing pedestrian motion prediction datasets benchmarks. As such, our contributions are three-fold:

\begin{enumerate}
    \item {Selection and engineering adaptation of a number of state-of-the-art pedestrian motion prediction methods to enable their online performance inside an autonomous driving framework.}
    \item {Creation of an experimental dataset from raw data recorded during past trips of the autonomous vehicle in Tartu, Estonia.}
    \item {Evaluation of the selected methods and baseline inside the Autoware-mini framework under different output modalities, creating unique perspective of state-of-the-art prediction methods performance.}
\end{enumerate}

\section{Research Methods}

\subsection{Problem Formulation}

We approach the pedestrian motion prediction problem primarily as the problem of future trajectory prediction based on historical observations, or, as they are sometimes referred, stimuli \cite{rudenko2020human}. There are several commonly used stimuli, however the most prominent is the previous locations of the pedestrian agents represented in 2D from a Birds-Eye-View (BEV) perspective \cite{schuetz2023review}. As such, we can formalize the problem of prediction in the following way: given the historical trajections of $N$ pedestrians over $H$ previous states $X_i = {x_i^1, x_i^2, ..., x_i^H}$, find the future trajectories of the pedestrians over $F$ future states $Y_i = {y_i^{H+1}, y_i^{H+2}, y_i^{H+F}}$, or in another form, find function $f$ that satisfies $Y_i = f(X_i)$ \cite{huang2022survey}. 

This simple problem formulation is used when introducing physics-based solutions, such as Constant Velocity Model (CVM) \cite{scholler2020constant}. However, it is insufficient for the more recent data-driven approaches. To extend it, we can represent $x_i^j$ as not only the physical location of the pedestrian $p_i^j$, but as a collection of location and other auxiliary inputs $x_i^j=(p_i^j, a_i^j, b_i^j, ...)$ such as environment representation (HD-map), inner characteristics of the pedestrians (gestures, emotions) or raw sensor data. Then, data-driven approaches need to estimate model $M$ that represents output trajectory $Y_i$ given input features of all $N$ pedestrians to represent cross-actor interactions: $Y_i=M(X_i,\{X_j\}_1^N)$. Often, instead of direct deterministic output, the model represents probability distribution $p(Y_i|X_i,\{X_j\}_1^N)$ from which multiple candidate trajectories can be sampled, with the intent that at least one of them will be close to future ground-truth.

\subsection{Motion prediction Algorithms}

The main step in the selection of state-of-the-art pedestrian motion prediction methods is finding them. The original search was performed in three distinct steps: first, the relevant keywords search was performed on academic journals to find corresponding survey articles \cite{rudenko2020human} \cite{huang2022survey} \cite{schuetz2023review} covering latest advancements in the field; second - filtering out articles of interest from all reviewed by reading the original publication and finding if they fulfill necessary criteria; third - analyzing the open source solution of algorithms and cross-referencing the published results with benchmarks of the used datasets. The criteria for filtering out the articles are as follows:

\begin{enumerate}
    \item The model is based on a supervised learning algorithm
    \item Pre-trained model allows real-time inference on the limited hardware constraints.
    \item The model's main input should be the historical trajectories of the pedestrians. While many models rely on other input data such as HD-map, in practice, it is hard to transfer learned features from the HD-map used for the training to the HD-map used in different autonomous stacks.
    \item Open source implementation of the model's architecture, preferably with access to pre-trained weights that reproduce results published in the article. While open-source implementation distributions are common, they are often incomplete, with details missing both from the training code and the description of the training process in the original publication.
\end{enumerate}

\begin{table}[!b]
\caption{Overview of chosen methods \label{tab:table1}}
\centering
\begin{tabular}{|c|c|c|c|c|}
\hline
Model & Year & Input & \makecell{Cross-agent \\ interaction \\ consideration} & \makecell{Output \\ Modality} \\
\hline
PECNet & 2020 & Trajectory & Yes & Stochastic \\
SGNet & 2022 & Trajectory & No & Stochastic \\
GATraj & 2023 & Trajectory & Capable & Probabilistic \\
MUSE & 2022 & Trajectory+Map & No & Probabilistic \\
\hline
CVM & - & Final Velocity & No & Deterministic \\
\hline
\end{tabular}
\end{table}

Following these criteria, four methods from the last four years were chosen: a) PECNet \cite{mangalam2020not} b) SGNet \cite{wang2022stepwise} c) GATraj \cite{cheng2023gatraj} d) MUSEVae \cite{lee2022muse}. Furthermore, we establish CVM as a useful baseline that is proven to be comparable and even outperforms data-driven approaches under certain conditions \cite{uhlemann2024evaluating}. The overhead comparison of methods according to our problem formulation is presented in Table \ref{tab:table1}. While we specified that reliance on HD-map is a negative criteria, MUSE-VAE method uses the semantic map as an auxiliary input with very few encoding categories that allow reliable adaptation of our existing mapping data. GATraj can integrate interactions between separate pedestrians, but the ablated version of the model (dropping GCN module) without this feature was used due to hardware constraints. Finally, stochastic output modality denotes that corresponding models output multiple predicted trajectories at once, however, models don't assign a numerical probability to output trajectories, while probabilistic models both output multiple trajectories and assign the probability, making it possible to choose the most likely trajectory(ies) for the evaluation.

\subsection{Experimental Setup}

\begin{figure*}[!t]
\centering
\vspace*{6pt}
\includegraphics[width=2.0\columnwidth]{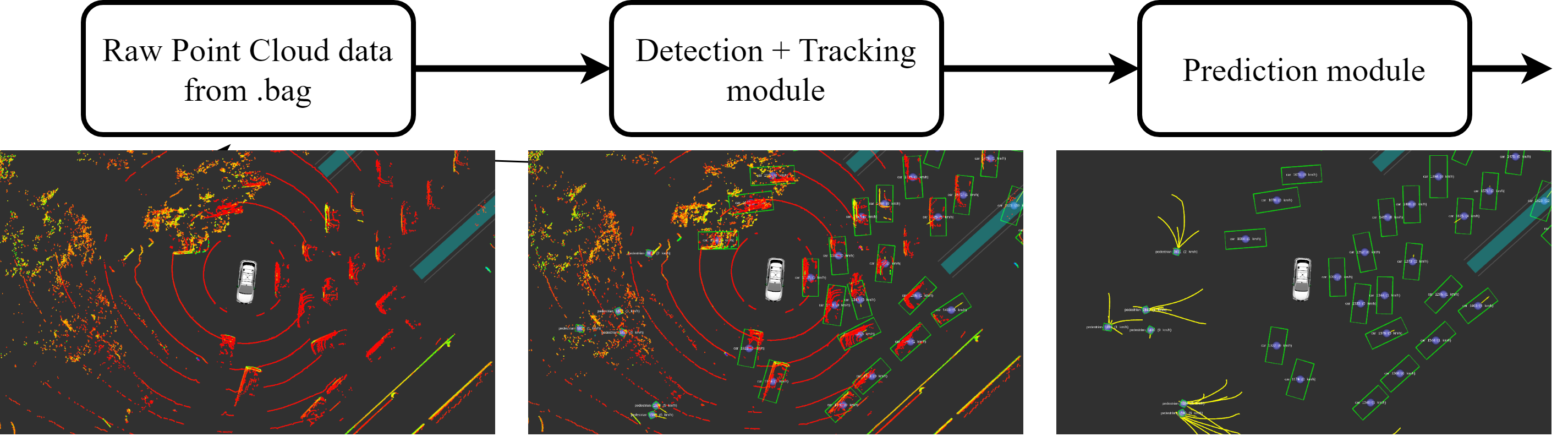}
\caption{Data flow inside Autoware Mini framework. Detection module extracts the shapes of the objects from raw point cloud, and classifies them to pedestrian/car/other objects. Afterwards, selected prediction model outputs candidate trajectories, represented here as yellow curves}
\label{fig_data_flow}
\end{figure*}

\subsubsection{Data collection}

Our goal is the evaluation of the existing prediction models in the environment as close as possible to a real autonomous vehicle. As such, we base our setup on Autoware Mini framework \cite{Matiisen_2023}, a lightweight fork written in pure Python of the original Autoware \cite{kato2018autoware}, one of the first modular autonomous driving frameworks, which itself is based on Robot Operating System (ROS) \cite{quigley2009ros}. This framework exists both as a scientific and a pedagogical tool, but it has also been tested on Lexus RX450h vehicle on the streets of Tartu, Estonia. The vehicle is equipped with Ouster OS1-128 and Velodyne VLP-32C lidars as the primary sensors for object detection. All data these and other sensors recorded during regular autonomous and manual trips for the last three years are stored inside .bag file format. This raw sensor data collection provides a primary dataset for our evaluation purposes. However, the full data collection is 3400+ different .bag files, totaling over 43 Tb, so some filtering was necessary to produce a compact and interesting dataset. The main filter was the rich presence of the pedestrians, requiring over five detected pedestrians to be present at the same time in every selected file and restricting the car route to the center of Tartu. The final result was a dataset with 18 .bag files, where every .bag file represents continuous scene, with a total runtime of 4 hours and 15 minutes of raw sensor footage.

\subsubsection{Implementation Details}

Using .bag files inside Autoware Mini framework allows to "replay" them, simulating the stream of data that the real autonomous vehicle received during the trip and allowing to run a full framework processing the data through full autonomous driving architecture described in Figure \ref{fig_architecture_gap}. This flow of data is represented in the Figure \ref{fig_data_flow}. Bag files output raw lidar point cloud data to the detector module. For detection, our framework uses SFA detector based on \cite{li2020rtm3d} to process raw point cloud data and output labelled vector representation of the objects. Next, the simplistic exponential moving averages based tracker \cite{hunter1986exponentially} keeps the objects permanence that allows to construct historic pedestrian trajectories in real-time. Finally, prediction module receives historical trajectories and other auxiliary data, and outputs candidate trajectories.

However, because of the errors introduced by these upstream algorithms, the raw dataset is unsuitable for chosen prediction models training, which necessitates usage of the pre-trained models. For our evaluation, we use PECNet and MUSE-VAE pre-trained on SDD \cite{robicquet2016learning}, while SGNet and GATraj are pre-trained on ETH/UCY  \cite{pellegrini2010improving, lerner2007crowds}. Both datasets follow the same input format of 8-point historical trajectories to predict 12-point candidate trajectory, where points are 0.4 seconds apart. As a result, all models consider 3.2 seconds of past data, to predict next 4.8 seconds of motion. Consequently, all models were adapted to run with a constant stream of incoming data, where every 0.4 seconds, they receive updated trajectories of all visible pedestrians. As such, the key requirement for performance was that the inference with a variable batch size finishes in 0.4 seconds before the next set of trajectories is obtained.

To simulate the limited resources of an embedded system of the real car, the full evaluation framework was performed on a consumer-grade laptop equipped with one NVIDIA GeForce 3080 and AMD Ryzen 5900HX. This severely limits performance compared to how these prediction models are usually evaluated - using either NVIDIA V100/A100, or multiple consumer-grade cards.

\begin{table*}[!t]
\centering
\vspace*{6pt}
\caption{Quantitative results with varying amount of candidate trajectories $k$ \label{tab:table2}}
\begin{tabular}{|c|c|c|c|c|c|c|c|c|c|c|c|c|c|}
\hline
Model & CVM & \multicolumn{3}{|c|}{PECNet} & \multicolumn{3}{|c|}{SGNet} & \multicolumn{3}{|c|}{GATraj} & \multicolumn{3}{|c|}{MUSE-VAE} \\
\hline
$k$ & 1 & 1 & 5 & 10 & 1 & 5 & 10 & 1 & 5 & 10 & 1 & 5 & 10 \\
\hline
minDynADE (m) & 1.605 & 1.749 & 1.637 & 1.564 & 1.845 & 1.390 & 1.297 & 1.871 & 1.133 & 0.918 & 2.587 & 1.909 & - \\
\hline
minDynFDE (m) & 2.970 & 3.276 & 3.021 & 2.854 & 3.476 & 2.649 & 2.458 & 3.589 & 2.012 & 1.563 & 5.212 & 3.739 & - \\
\hline
\end{tabular}
\end{table*}

\subsection{Evaluation Metrics}

The most commonly used metrics for evaluating trajectory prediction are Average Displacement Error (ADE) and Final Displacement Error (FDE) \cite{rudenko2020human}. ADE metric is commonly defined as the average L2 distance between all points of predicted trajectory and ground truth future trajectory:
$$ADE = \frac{1}{N}\frac{1}{F}\sum_{i=1}^{N}\sum_{j=1}^{F}||y_i^{H+j} - \hat{y}_i^{H+j}||_2$$, where $\hat{Y}_i = {\hat{y}_i^{H+1}, \hat{y}_i^{H+2}, \hat{y}_i^{H+F}}$ is the ground truth future trajectory. FDE is similar, but considers only L2 distance between final points of predicted and ground-truth trajectories. $$FDE = \frac{1}{N}\sum_{i=1}^{N}||y_i^{H+F} - \hat{y}_i^{H+F}||_2$$

However, many data-driven approaches output multiple candidate trajectories. To solve this, the variations of ADE/FDE were introduced called minADE/minFDE, which calculates metrics for every candidate trajectory but considers only the best (lowest) metric score. This variant is commonly used for benchmarking on most pedestrian motion prediction datasets, such as SDD, ETH-UCY, Nuscenes, and Argoverse, with the only difference being that dataset evaluation restricts the amount of candidate trajectories methods that can be submitted. Early datasets such as SDD and ETH-UCY allow up to $k=20$ candidate trajectories, while Nuscenes has two different benchmarks for $k=5$ and $k=10$ and Argoverse v1/2 settling down for $k=6$. 

The methodology of using only best score (Best-of-\emph{N} approach) has been recently called into question \cite{thiede2019analyzing, ivanovic2021rethinking, wu2023truly, mohamed2022social, shahroudievaluation}. The central point of the critique, is that if the future ground truth trajectories are parametric distribution (Gaussian or mixed-Gaussian in the simplest case), and our model tries to estimate this distribution by means of minADE/minFDE, it instead estimates square root of PDF \cite{thiede2019analyzing}, which means that it will always perform worse than Bayes-optimal predictor. The theoretical way to recover from this is significantly increase amount of $k$ trajectories we sample. However, this approach is unpractical in autonomous driving, where there are performance constraints to obtaining more sample trajectories if our inference time is long, and a large amount of predicted trajectories can cause a problem for downstream planning and control modules (when for example pedestrian is moving on the sidewalk, and low probability candidate trajectory falsely predicts they will cross a road, forcing vehicle to react, when with lower amount of candidate trajectories this prediction wouldn't exist). Another point of the criticism is that existing metrics poorly represent the so-called "dynamics gap" \cite{wu2023truly} - the idea is that as much as our autonomous vehicle planner considers predicted trajectories and reacts to them, the agents whose motion we predict also consider and predict our ego-vehicle movements, creating dependency on each others' predictions. To resolve this, we follow \cite{wu2023truly}, and implement Dynamic Average Displacement Error (DynADE) and Dynamic Final Displacement Error (DynFDE) metrics. The critical difference is the interactivity factor - in normal ADE/FDE calculation, we calculate metric once per given trajectory in the dataset, while in our setup as we receive updated trajectory of each agent in online stream of data, we calculate new metric value with every new added trajectory point, averaging these L2 distance values for every agent, and then averaging over all agents in the end of each scene. Formally:

$$DynADE = \frac{1}{N}\sum_{i=1}^{N}\frac{1}{L_i}\sum_{j=1}^{L_i}\frac{1}{F}\sum_{k=1}^{F}||y_i^{j+k} - \hat{y}_i^{j+k}||_2$$
$$DynFDE = \frac{1}{N}\sum_{i=1}^{N}\frac{1}{L_i}\sum_{j=1}^{L_i}||y_i^{j+F} - \hat{y}_i^{j+F}||_2$$, 
where $L_i$ is the total length of agent $i$ trajectory in the observed scene. For our dataset we average results from each .bag file scene to obtain the final result. The new metric is similar to standard ADE/FDE but shows a more significant correlation with driving performance when integrated into the autonomous driving framework \cite{wu2023truly}. Correspondingly, for models that output multiple candidate trajectories, Best-of-\emph{N} approach is still used by choosing the candidate trajectory with the lowest metric score to allow comparison between performance on our dataset and standard datasets such as Nuscenes, denoted as minDynADE/minDynFDE.

\section{Findings}

\subsection{Quantitative results}

We first present a quantitative evaluation of the chosen algorithms and baseline over a created dataset in Table \ref{tab:table2}. To demonstrate the effect of the Best-of-\emph{N} approach, we test selected methods under variable candidate trajectories amount $k=1,5,10$. If the model provides more trajectories than $k$ during the single inference run, then $k$ trajectories were selected randomly unless the specified method also outputs the probability of every candidate trajectory, in which case $k$ most likely trajectories were selected. As previously stated, all methods except baseline CVM rely on trajectories over the past 3.2 seconds while outputting candidate trajectories for 4.8 seconds, in line with pre-trained model original datasets ETH/UCY and SDD. Several insights can be derived from the results:

\begin{itemize}
    \item Inline with theory, Best-of-\emph{N} metric of every data-driven model increases with increase of $k$. However, increasing $k$ beyond 10 is impractical in the framework due to performance limitations and potential issues in downstream modules such as planning.
    \item Given $k=1$, none of the models outperform the Constant Velocity Model despite considering more than a final state of the pedestrian trajectory. This confirms previous findings \cite{uhlemann2024evaluating} and demonstrates that pedestrian motion prediction remains an unsolved problem under certain conditions, opening avenues for future research that need to consider other input modalities instead of relying on trajectory only.
    \item MUSE-VAE model, despite being the only model that relies on additional auxiliary map input data, performs the weakest in our framework. The result shows that transferring learned features from one dataset to another is non-trivial, even for the simple semantic map representation. Additionally, MUSE-VAE implementation wasn't able to output $k=10$ candidate trajectories under the required time constraint of 0.4 seconds, leading to no results available for Table \ref{tab:table2}.
\end{itemize}

% \subsection{Qualitative results}
% Potential section to add, however space limitations

\begin{table*}[!t]
\centering
\vspace*{6pt}
\caption{Ablation testing with varying amount of considered historical steps $H$ \label{tab:table3}}
\begin{tabular}{|c|c|c|c|c|c|c|c|c|c|c|c|c|c|}
\hline
Model & CVM & \multicolumn{3}{|c|}{PECNet} & \multicolumn{3}{|c|}{SGNet} & \multicolumn{3}{|c|}{GATraj} & \multicolumn{3}{|c|}{MUSE-VAE} \\
\hline
$H$ & 0 & 2 & 4 & 8 & 2 & 4 & 8 & 2 & 4 & 8 & 2 & 4 & 8 \\
\hline
minDynADE (m) & 1.605 & 2.043 & 1.824 & 1.637 & 2.194 & 1.479 & 1.390 & 1.281 & 1.165 & 1.133 & 1.910 & 1.964 & 1.909 \\
\hline
minDynFDE (m) & 2.970 & 3.762 & 3.382 & 3.021 & 3.661 & 2.807 & 2.649 & 2.268 & 2.049 & 2.012 & 3.862 & 3.861 & 3.739 \\
\hline
\end{tabular}
\end{table*}

\subsection{Ablation testing}

The critical issue for pedestrian motion prediction in the context of autonomous driving is the limited availability of historical information. During testing on our dataset, it is common that some pedestrians are only detected very close to the ego-vehicle, giving no time to construct a full 3.2-second trajectory history before the critical prediction needs to be made. As such, we perform ablation testing where we limit the amount of historical trajectory points models consider $H$ to lower value, simulating the lower length of historical trajectory, and compare with the results on default behavior $H=8$ in table \ref{tab:table3} (for all methods except CVM $k=5$). From this testing, it can be seen that while PECNet gradually improves its predictions while incorporating more historical trajectory steps, GATraj and SGNet plateau performance on $H=4$, and Muse-VAE seems to rely almost entirely on auxiliary map information and last $H=2$ trajectory points, as no meaningful improvements observed with increase in $H$. The testing shows that while models claim that they consider entire past trajectories, they do not always encode useful information from the oldest trajectory steps.

\subsection{A note on reproducibility of experiments}

Due to the internal workings of ROS adding delays during data transfer between different framework modules, it is impossible to guarantee deterministic metrics calculation on a specific .bag scene. As such, we execute one scene evaluation 20 times using every predictor (at $k=5$ and $H=8$) and calculate the standard deviation and mean of the obtained metric values to showcase possible deviation ranges from the published results in this section. The results are available in Table \ref{tab:table4}

\begin{table}[!h]
\caption{Possible values of standard deviation during reproduction of the results \label{tab:table4}}
\centering
\begin{tabular}{|c|c|c|c|c|}
\hline
Model & \multicolumn{2}{|c|}{MinDynADE} & \multicolumn{2}{|c|}{MinDynFDE} \\
& Mean & Std & Mean & Std \\
\hline
PECNet &  1.427 & 0.049 & 2.615 &  0.096 \\
SGNet &  1.081 & 0.061 &  2.056 & 0.118 \\
GATraj & 0.935 & 0.044 & 1.603 & 0.084 \\
MUSE & 1.484 & 0.074 & 2.808 & 0.136 \\
\hline
CVM & 1.451 & 0.051 & 2.724 & 0.090 \\
\hline
\end{tabular}
\end{table}

\section{Conclusion}

In this work, we have performed a comprehensive evaluation of selected state-of-the-art pedestrian motion prediction methods on the dataset obtained from a real autonomous vehicle trips in the software environment of the autonomous driving framework under similar hardware restraints. Our evaluation focused on testing limits of the Best-of-\emph{N} approach of the standard ADE/FDE metrics using modified implementation, created to resolve issues arising from implementing standard solutions into full modular autonomous driving stack. We found out that none of the evaluated models perform better than the standard Constant Velocity Model when outputing only a single candidate trajectory. Otherwise, GATraj model performs the best, both when increasing amount of candidate trajectories beyond one, and when decreasing the length of the observation window of historical trajectory input.

The study confirms that pedestrian motion prediction remains an open problem under certain condition, and that researchers should more strictly evaluate potential practical application of their research. Many better or more promising machine-learning models were not evaluated in this study due to hardware or implementation constraints. As such, future research direction should be focused towards more light-weight practical solutions to the problem, while also developing methods of incorporating input data other than historical trajectories. We are hopeful that our research is helpful both to scientists working on new pedestrian motion prediction models and engineers working on real autonomous driving vehicles.

\section*{Acknowledgments}

This work was supported in part by European Social Fund through the "ICT Programme" Measure and in part by Bolt Technologies through the Collaboration Project under Grant LLTAT21278.

\bibliographystyle{IEEEtran}
\bibliography{bibtex/bib/IEEEabrv.bib,bibtex/bib/main.bib}

% Generated by IEEEtran.bst, version: 1.14 (2015/08/26)
\begin{thebibliography}{10}
\providecommand{\url}[1]{#1}
\csname url@samestyle\endcsname
\providecommand{\newblock}{\relax}
\providecommand{\bibinfo}[2]{#2}
\providecommand{\BIBentrySTDinterwordspacing}{\spaceskip=0pt\relax}
\providecommand{\BIBentryALTinterwordstretchfactor}{4}
\providecommand{\BIBentryALTinterwordspacing}{\spaceskip=\fontdimen2\font plus
\BIBentryALTinterwordstretchfactor\fontdimen3\font minus \fontdimen4\font\relax}
\providecommand{\BIBforeignlanguage}[2]{{%
\expandafter\ifx\csname l@#1\endcsname\relax
\typeout{** WARNING: IEEEtran.bst: No hyphenation pattern has been}%
\typeout{** loaded for the language `#1'. Using the pattern for}%
\typeout{** the default language instead.}%
\else
\language=\csname l@#1\endcsname
\fi
#2}}
\providecommand{\BIBdecl}{\relax}
\BIBdecl

\bibitem{herrmann2018autonomous}
A.~Herrmann, W.~Brenner, and R.~Stadler, \emph{Autonomous driving: how the driverless revolution will change the world}.\hskip 1em plus 0.5em minus 0.4em\relax Emerald Group Publishing, 2018.

\bibitem{californiadmv_2021}
\BIBentryALTinterwordspacing
``Dmv approves cruise and waymo to use autonomous vehicles for commercial service in designated parts of bay area,'' Sep 2021. [Online]. Available: \url{https://www.dmv.ca.gov/portal/news-and-media/117199-2/}
\BIBentrySTDinterwordspacing

\bibitem{rudenko2020human}
A.~Rudenko, L.~Palmieri, M.~Herman, K.~M. Kitani, D.~M. Gavrila, and K.~O. Arras, ``Human motion trajectory prediction: A survey,'' \emph{The International Journal of Robotics Research}, vol.~39, no.~8, pp. 895--935, 2020.

\bibitem{zegeer2012pedestrian}
C.~V. Zegeer and M.~Bushell, ``Pedestrian crash trends and potential countermeasures from around the world,'' \emph{Accident Analysis \& Prevention}, vol.~44, no.~1, pp. 3--11, 2012.

\bibitem{branche2008world}
C.~Branche, J.~Ozanne-Smith, K.~Oyebite, and A.~A. Hyder, ``World report on child injury prevention,'' 2008.

\bibitem{scholler2020constant}
C.~Sch{\"o}ller, V.~Aravantinos, F.~Lay, and A.~Knoll, ``What the constant velocity model can teach us about pedestrian motion prediction,'' \emph{IEEE Robotics and Automation Letters}, vol.~5, no.~2, pp. 1696--1703, 2020.

\bibitem{schuetz2023review}
E.~Schuetz and F.~B. Flohr, ``A review of trajectory prediction methods for the vulnerable road user,'' \emph{Robotics}, vol.~13, no.~1, p.~1, 2023.

\bibitem{lee2022muse}
M.~Lee, S.~S. Sohn, S.~Moon, S.~Yoon, M.~Kapadia, and V.~Pavlovic, ``Muse-vae: Multi-scale vae for environment-aware long term trajectory prediction,'' in \emph{Proceedings of the IEEE/CVF conference on computer vision and pattern recognition}, 2022, pp. 2221--2230.

\bibitem{zhou2023csr}
H.~Zhou, D.~Ren, X.~Yang, M.~Fan, and H.~Huang, ``Csr: cascade conditional variational auto encoder with socially-aware regression for pedestrian trajectory prediction,'' \emph{Pattern Recognition}, vol. 133, p. 109030, 2023.

\bibitem{wang2022stepwise}
C.~Wang, Y.~Wang, M.~Xu, and D.~J. Crandall, ``Stepwise goal-driven networks for trajectory prediction,'' \emph{IEEE Robotics and Automation Letters}, vol.~7, no.~2, pp. 2716--2723, 2022.

\bibitem{chen2022scept}
Y.~Chen, B.~Ivanovic, and M.~Pavone, ``Scept: Scene-consistent, policy-based trajectory predictions for planning,'' in \emph{Proceedings of the IEEE/CVF conference on computer vision and pattern recognition}, 2022, pp. 17\,103--17\,112.

\bibitem{li2021temporal}
Y.~Li, R.~Liang, W.~Wei, W.~Wang, J.~Zhou, and X.~Li, ``Temporal pyramid network with spatial-temporal attention for pedestrian trajectory prediction,'' \emph{IEEE Transactions on Network Science and Engineering}, vol.~9, no.~3, pp. 1006--1019, 2021.

\bibitem{zhou2022gchgat}
L.~Zhou, Y.~Zhao, D.~Yang, and J.~Liu, ``Gchgat: Pedestrian trajectory prediction using group constrained hierarchical graph attention networks,'' \emph{Applied Intelligence}, vol.~52, no.~10, pp. 11\,434--11\,447, 2022.

\bibitem{yang2023social}
C.~Yang, H.~Pan, W.~Sun, and H.~Gao, ``Social self-attention generative adversarial networks for human trajectory prediction,'' \emph{IEEE Transactions on Artificial Intelligence}, 2023.

\bibitem{ngiam2021scene}
J.~Ngiam, B.~Caine, V.~Vasudevan, Z.~Zhang, H.-T.~L. Chiang, J.~Ling, R.~Roelofs, A.~Bewley, C.~Liu, A.~Venugopal \emph{et~al.}, ``Scene transformer: A unified architecture for predicting multiple agent trajectories,'' \emph{arXiv preprint arXiv:2106.08417}, 2021.

\bibitem{nayakanti2023wayformer}
N.~Nayakanti, R.~Al-Rfou, A.~Zhou, K.~Goel, K.~S. Refaat, and B.~Sapp, ``Wayformer: Motion forecasting via simple \& efficient attention networks,'' in \emph{2023 IEEE International Conference on Robotics and Automation (ICRA)}.\hskip 1em plus 0.5em minus 0.4em\relax IEEE, 2023, pp. 2980--2987.

\bibitem{cheng2023gatraj}
H.~Cheng, M.~Liu, L.~Chen, H.~Broszio, M.~Sester, and M.~Y. Yang, ``Gatraj: A graph-and attention-based multi-agent trajectory prediction model,'' \emph{ISPRS Journal of Photogrammetry and Remote Sensing}, vol. 205, pp. 163--175, 2023.

\bibitem{mao2023leapfrog}
W.~Mao, C.~Xu, Q.~Zhu, S.~Chen, and Y.~Wang, ``Leapfrog diffusion model for stochastic trajectory prediction,'' in \emph{Proceedings of the IEEE/CVF conference on computer vision and pattern recognition}, 2023, pp. 5517--5526.

\bibitem{bae2024singulartrajectory}
I.~Bae, Y.-J. Park, and H.-G. Jeon, ``Singulartrajectory: Universal trajectory predictor using diffusion model,'' in \emph{Proceedings of the IEEE/CVF Conference on Computer Vision and Pattern Recognition}, 2024, pp. 17\,890--17\,901.

\bibitem{keysan2023can}
A.~Keysan, A.~Look, E.~Kosman, G.~G{\"u}rsun, J.~Wagner, Y.~Yu, and B.~Rakitsch, ``Can you text what is happening? integrating pre-trained language encoders into trajectory prediction models for autonomous driving,'' \emph{arXiv preprint arXiv:2309.05282}, 2023.

\bibitem{bae2024can}
I.~Bae, J.~Lee, and H.-G. Jeon, ``Can language beat numerical regression? language-based multimodal trajectory prediction,'' in \emph{Proceedings of the IEEE/CVF Conference on Computer Vision and Pattern Recognition}, 2024, pp. 753--766.

\bibitem{ivanovic2021rethinking}
B.~Ivanovic and M.~Pavone, ``Rethinking trajectory forecasting evaluation,'' \emph{arXiv preprint arXiv:2107.10297}, 2021.

\bibitem{wu2023truly}
H.~Wu, T.~Phong, C.~Yu, P.~Cai, S.~Zheng, and D.~Hsu, ``What truly matters in trajectory prediction for autonomous driving?'' \emph{arXiv preprint arXiv:2306.15136}, 2023.

\bibitem{shridhar2021beelines}
S.~Shridhar, Y.~Ma, T.~Stentz, Z.~Shen, G.~C. Haynes, and N.~Traft, ``Beelines: Motion prediction metrics for self-driving safety and comfort,'' in \emph{2021 IEEE International Conference on Robotics and Automation (ICRA)}.\hskip 1em plus 0.5em minus 0.4em\relax IEEE, 2021, pp. 881--887.

\bibitem{mohamed2022social}
A.~Mohamed, D.~Zhu, W.~Vu, M.~Elhoseiny, and C.~Claudel, ``Social-implicit: Rethinking trajectory prediction evaluation and the effectiveness of implicit maximum likelihood estimation,'' in \emph{European Conference on Computer Vision}.\hskip 1em plus 0.5em minus 0.4em\relax Springer, 2022, pp. 463--479.

\bibitem{pellegrini2010improving}
S.~Pellegrini, A.~Ess, and L.~Van~Gool, ``Improving data association by joint modeling of pedestrian trajectories and groupings,'' in \emph{Computer Vision--ECCV 2010: 11th European Conference on Computer Vision, Heraklion, Crete, Greece, September 5-11, 2010, Proceedings, Part I 11}.\hskip 1em plus 0.5em minus 0.4em\relax Springer, 2010, pp. 452--465.

\bibitem{lerner2007crowds}
A.~Lerner, Y.~Chrysanthou, and D.~Lischinski, ``Crowds by example,'' in \emph{Computer graphics forum}, vol.~26, no.~3.\hskip 1em plus 0.5em minus 0.4em\relax Wiley Online Library, 2007, pp. 655--664.

\bibitem{robicquet2016learning}
A.~Robicquet, A.~Sadeghian, A.~Alahi, and S.~Savarese, ``Learning social etiquette: Human trajectory understanding in crowded scenes,'' in \emph{Computer Vision--ECCV 2016: 14th European Conference, Amsterdam, The Netherlands, October 11-14, 2016, Proceedings, Part VIII 14}.\hskip 1em plus 0.5em minus 0.4em\relax Springer, 2016, pp. 549--565.

\bibitem{chang2019argoverse}
M.-F. Chang, J.~Lambert, P.~Sangkloy, J.~Singh, S.~Bak, A.~Hartnett, D.~Wang, P.~Carr, S.~Lucey, D.~Ramanan \emph{et~al.}, ``Argoverse: 3d tracking and forecasting with rich maps,'' in \emph{Proceedings of the IEEE/CVF conference on computer vision and pattern recognition}, 2019, pp. 8748--8757.

\bibitem{ettinger2021large}
S.~Ettinger, S.~Cheng, B.~Caine, C.~Liu, H.~Zhao, S.~Pradhan, Y.~Chai, B.~Sapp, C.~R. Qi, Y.~Zhou \emph{et~al.}, ``Large scale interactive motion forecasting for autonomous driving: The waymo open motion dataset,'' in \emph{Proceedings of the IEEE/CVF International Conference on Computer Vision}, 2021, pp. 9710--9719.

\bibitem{huang2022survey}
Y.~Huang, J.~Du, Z.~Yang, Z.~Zhou, L.~Zhang, and H.~Chen, ``A survey on trajectory-prediction methods for autonomous driving,'' \emph{IEEE Transactions on Intelligent Vehicles}, vol.~7, no.~3, pp. 652--674, 2022.

\bibitem{mangalam2020not}
K.~Mangalam, H.~Girase, S.~Agarwal, K.-H. Lee, E.~Adeli, J.~Malik, and A.~Gaidon, ``It is not the journey but the destination: Endpoint conditioned trajectory prediction,'' in \emph{Computer Vision--ECCV 2020: 16th European Conference, Glasgow, UK, August 23--28, 2020, Proceedings, Part II 16}.\hskip 1em plus 0.5em minus 0.4em\relax Springer, 2020, pp. 759--776.

\bibitem{uhlemann2024evaluating}
N.~Uhlemann, F.~Fent, and M.~Lienkamp, ``Evaluating pedestrian trajectory prediction methods with respect to autonomous driving,'' \emph{IEEE Transactions on Intelligent Transportation Systems}, 2024.

\bibitem{Matiisen_2023}
\BIBentryALTinterwordspacing
T.~Matiisen, ``Ut-adl/autoware\_mini: Autoware mini is a minimalistic python-based autonomy software.'' 2023. [Online]. Available: \url{https://github.com/UT-ADL/autoware_mini/}
\BIBentrySTDinterwordspacing

\bibitem{kato2018autoware}
S.~Kato, S.~Tokunaga, Y.~Maruyama, S.~Maeda, M.~Hirabayashi, Y.~Kitsukawa, A.~Monrroy, T.~Ando, Y.~Fujii, and T.~Azumi, ``Autoware on board: Enabling autonomous vehicles with embedded systems,'' in \emph{2018 ACM/IEEE 9th International Conference on Cyber-Physical Systems (ICCPS)}.\hskip 1em plus 0.5em minus 0.4em\relax IEEE, 2018, pp. 287--296.

\bibitem{quigley2009ros}
M.~Quigley, K.~Conley, B.~Gerkey, J.~Faust, T.~Foote, J.~Leibs, R.~Wheeler, A.~Y. Ng \emph{et~al.}, ``Ros: an open-source robot operating system,'' in \emph{ICRA workshop on open source software}, vol.~3, no. 3.2.\hskip 1em plus 0.5em minus 0.4em\relax Kobe, Japan, 2009, p.~5.

\bibitem{li2020rtm3d}
P.~Li, H.~Zhao, P.~Liu, and F.~Cao, ``Rtm3d: Real-time monocular 3d detection from object keypoints for autonomous driving,'' in \emph{European Conference on Computer Vision}.\hskip 1em plus 0.5em minus 0.4em\relax Springer, 2020, pp. 644--660.

\bibitem{hunter1986exponentially}
J.~S. Hunter, ``The exponentially weighted moving average,'' \emph{Journal of quality technology}, vol.~18, no.~4, pp. 203--210, 1986.

\bibitem{thiede2019analyzing}
L.~A. Thiede and P.~P. Brahma, ``Analyzing the variety loss in the context of probabilistic trajectory prediction,'' in \emph{Proceedings of the IEEE/CVF International Conference on Computer Vision}, 2019, pp. 9954--9963.

\bibitem{shahroudievaluation}
N.~Shahroudi, M.~Lepson, and M.~Kull, ``Evaluation of trajectory distribution predictions with energy score,'' in \emph{Forty-first International Conference on Machine Learning}, 2024.

\end{thebibliography}

\end{document}